\documentclass{isprs} %
\usepackage{subfigure}
\usepackage{setspace}
\usepackage{geometry} %
\usepackage{epstopdf}
\usepackage[labelsep=period]{caption}  %
\usepackage[british]{babel} 
\usepackage[hang]{footmisc}
\usepackage{balance} %

\usepackage[authoryear]{natbib}

\renewcommand{\cite}{\citep}

\usepackage{bbm}
\usepackage{bm}
\usepackage{multirow}
\usepackage{booktabs}
\usepackage{amsmath}
\usepackage{amsfonts}
\usepackage{cleveref}
\usepackage{tikz}
\usetikzlibrary{calc}
\usetikzlibrary{positioning}
\usetikzlibrary{arrows.meta}
\usepackage{xcolor}
\usepackage{siunitx}
\usepackage{enumitem}

\renewcommand{\vec}[1]{\ensuremath{\mathbf{#1}}}
\newcommand{\vecs}[1]{\ensuremath{\bm{#1}}}

\usepackage{amssymb}
\usepackage{utfsym}
\newcommand{\crossmark}{\scalebox{0.75}{\usym{2613}}}

\newcommand{\goldmedal}{}
\newcommand{\silvermedal}{}

\newcommand{\firstplace}[1]{\textbf{#1}}
\newcommand{\secondplace}[1]{\underline{#1}}

\newcommand{\improvement}[1]{\tiny\textcolor{green!70!black}{#1\%\,$\downarrow$}}
\newcommand{\decrease}[1]{\tiny\textcolor{red!70!black}{#1\%\,$\downarrow$}}

\newcommand{\model}{SatGeo-NeRF}
\newcommand{\modelname}{\emph{\model{}}}

\usepackage{hyphenat}
\begin{document}
\title{SatGeo-NeRF: Geometrically Regularized NeRF for Satellite Imagery}
\date{}

 \author{
  Valentin Wagner\textsuperscript{1}, Sebastian Bullinger\textsuperscript{1}, Michael Arens\textsuperscript{1}, Rainer Stiefelhagen\textsuperscript{2} }

\address{
	\textsuperscript{1 }Fraunhofer IOSB, Ettlingen, Germany - \{\textit{firstname.lastname}\}@iosb.fraunhofer.de\\
	\textsuperscript{2 }Karlsruhe Institute of Technology, Karlsruhe, Germany - rainer.stiefelhagen@kit.edu
}

\abstract{

We present \modelname, a geometrically regularized \emph{NeRF} for satellite imagery that mitigates overfitting-induced geometric artifacts observed in current state-of-the-art models using three model-agnostic regularizers.
\emph{Gravity-Aligned Planarity Regularization} aligns depth-inferred, approximated surface normals with the gravity axis to promote local planarity, coupling adjacent rays via a corresponding surface approximation to facilitate cross-ray gradient flow.
\emph{Granularity Regularization} enforces a coarse\hyp{}to\hyp{}fine geometry\hyp{}learning scheme, and \emph{Depth\hyp{}Supervised Regularization} stabilizes early training for improved geometric accuracy.
On the DFC2019 satellite reconstruction benchmark, \modelname{} improves the Mean Altitude Error by 14.0\% and 11.4\% relative to state-of-the-art baselines such as \emph{EO-NeRF} and \emph{EO-GS}.

}


\maketitle

\section{Introduction}
\label{sec:intro}

In recent years, the number of earth observation satellites featuring high-resolution camera systems has increased drastically. 
While 3D information from satellite data is highly impactful for urban, environmental, and disaster domains, reliable reconstruction remains an open research question because the inverse problem is fundamentally ambiguous, hindering a direct mapping from 2D projections to 3D structure.

Satellite images pose domain-specific challenges including a)\ specialized camera models due to the vast distances involved, b)\ images captured over multiple satellite passes, resulting in variable shadow and lighting conditions, and c)\ temporary objects such as vehicles moving in between image captures.

Popular photogrammetry approaches such as \citet{s2p}, \citet{ames} and \citet{vissat} focus on image-based feature matching to extract explicit geometry representations such as point clouds.
Recent works \citep{snerf,satnerf,eonerf,sundial} re-approach the problem as a novel-view synthesis task using adaptations of \emph{Neural Radiance Fields} (\emph{NeRF}) \cite{nerf}. \emph{NeRF} reconstruct images on a per-pixel basis from multiple viewpoints, learning an implicit geometrical scene understanding as byproduct. 
The learned geometry is extracted into a \emph{Digital Surface Model} (\emph{DSM}) to quantitative evaluate the altitude error of the derived results.

\emph{NeRFs} build upon the multi-view consistency assumption, effectively expecting the appearance of the geometry to be consistent across views. Thus, \emph{NeRFs} optimize a pure photometric consistency at pixel level and impose no explicit geometric constraints on the scene.
This view-consistency assumption breaks under the high variability of multi-temporal satellite data.
Therefore, previous works model variable elements such as lighting as part of the rendering process \cite{snerf,eonerf,sundial} and introduce uncertainty to handle transient objects \cite{satnerf}.
The geometry is hereby still unconstrained. 

\begin{figure}[!t]
	\centering
	\begin{tikzpicture}[every node/.style={inner sep=0, outer sep=0}]
		\newcommand{\fullw}{0.47\linewidth}  %
		\newcommand{\cropw}{0.96\linewidth}  %
		\newcommand{\lw}{1.2pt}
		
		\colorlet{roiColor}{black}
		\definecolor{lightgray}{HTML}{FFFFE0} %
		\colorlet{roi2Color}{black}
		
		\tikzset{hl/.style={line width=\lw, draw=roiColor}}
		\tikzset{hl2/.style={line width=\lw, draw=roi2Color}}
		
		\node (A) at (0,0) {\includegraphics[width=\fullw]{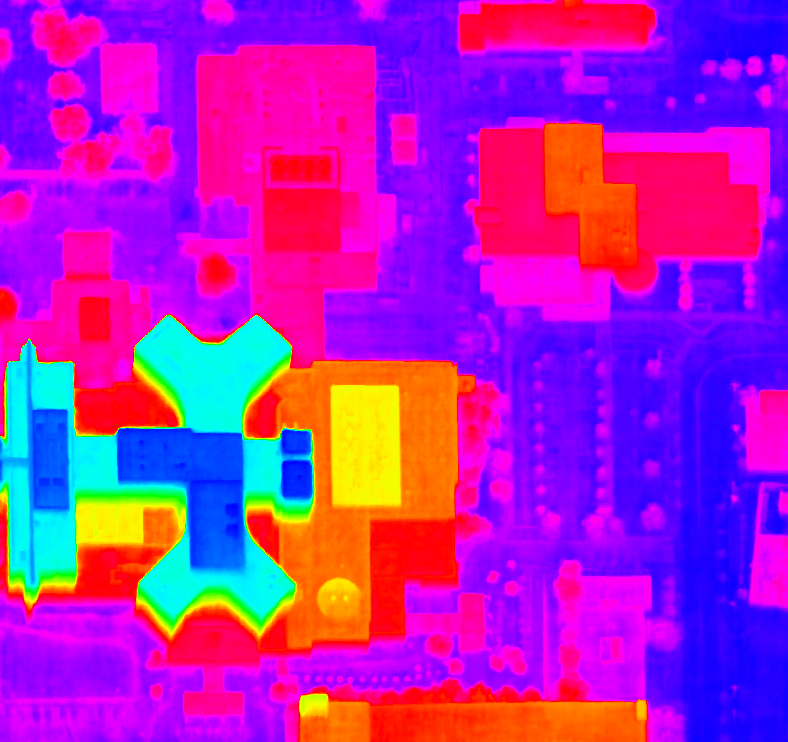}};
		\node[right=2mm of A] (B) {\includegraphics[width=\fullw]{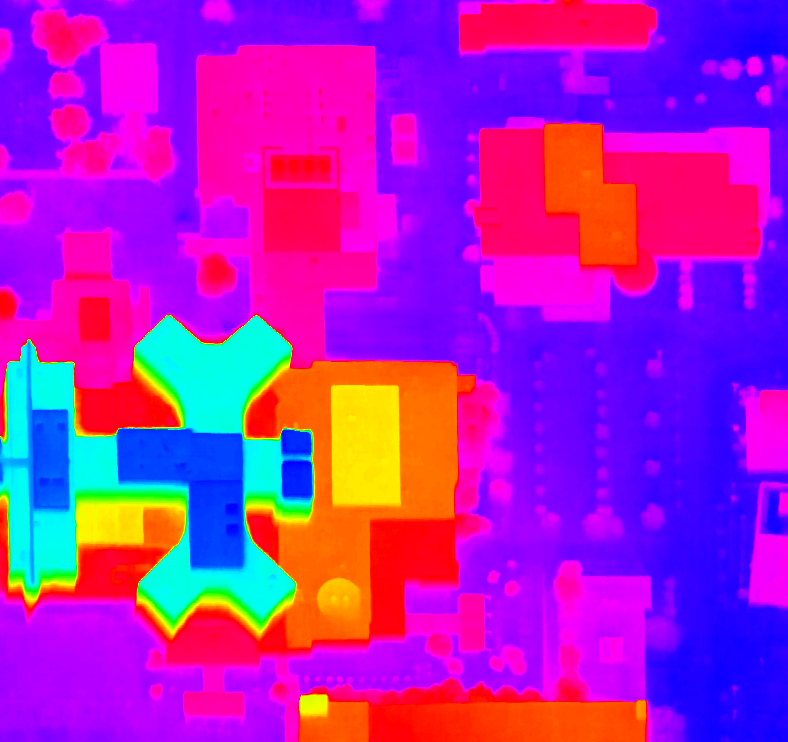}};
		
		\node[above=1mm of A, anchor=south] {EO-NeRF \cite{eonerf}};
		\node[above=1mm of B, anchor=south] {\model{} (Ours)};
		
		\def\Ax{0.01cm}\def\Ay{2.69cm}\def\Aw{3.84cm}\def\Ah{0.94cm}
		\coordinate (aNW) at ($(A.north west)+(\Ax,-\Ay)$);
		\coordinate (aSE) at ($(A.north west)+(\Ax+\Aw,-\Ay-\Ah)$);
		\coordinate (aSW) at (aNW |- aSE);
		\coordinate (aNE) at (aSE |- aNW);
		
		\def\Bx{\Ax}\def\By{\Ay}\def\Bw{\Aw}\def\Bh{\Ah}
		\coordinate (bNW) at ($(B.north west)+(\Bx,-\By)$);
		\coordinate (bSE) at ($(B.north west)+(\Bx+\Bw,-\By-\Bh)$);
		\coordinate (bSW) at (bNW |- bSE);
		\coordinate (bNE) at (bSE |- bNW);
		
		\draw[hl] (aNW) rectangle (aSE);
		\draw[hl2] (bNW) rectangle (bSE);
		
		\node[below=2mm of A.south west, anchor=north west] (Az)
		{\includegraphics[width=\cropw]{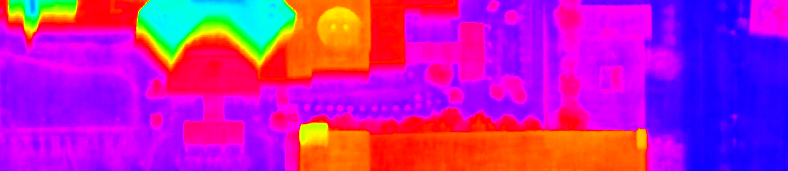}};
		\node[below=2mm of Az.south west, anchor=north west] (Bz)
		{\includegraphics[width=\cropw]{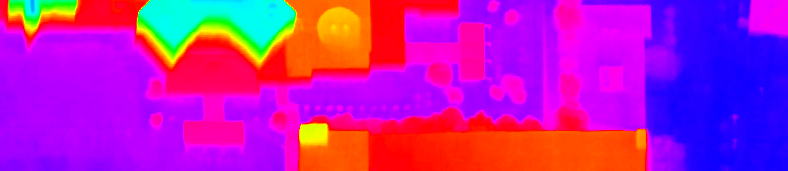}};
		\node[below=2mm of Bz.south west, anchor=north west] (Gz)
		{\includegraphics[width=\cropw]{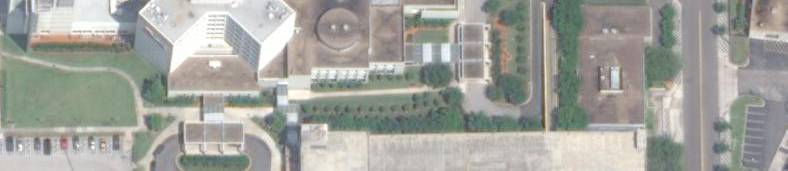}};
		\node[below=2mm of Gz.south west, anchor=north west] (Lz)
		{\includegraphics[width=\cropw]{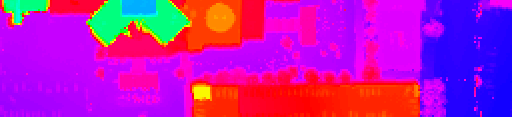}};
		
		\node[anchor=north west, fill=lightgray, draw=black, line width=0.8pt,
		inner sep=1pt, font=\bfseries\footnotesize]
		at ($(aNW.north west)+(1mm,-1mm)$) {A};
		\node[anchor=north west, fill=lightgray, draw=black, line width=0.8pt,
		inner sep=1pt, font=\bfseries\footnotesize]
		at ($(Az.north west)+(1mm,-1mm)$) {A};
		\node[anchor=north west, fill=lightgray, draw=black, line width=0.8pt,
		inner sep=1pt, font=\bfseries\footnotesize]
		at ($(Bz.north east)+(-3.5mm,-1mm)$) {B};
		\node[anchor=north west, fill=lightgray, draw=black, line width=0.8pt,
		inner sep=1pt, font=\bfseries\footnotesize]
		at ($(bNE.north west)+(-3.5mm,-1mm)$) {B};
		
		\draw[-{Latex[length=2.5mm]}, line width=2pt]
		($(Az.south west)+(1.4cm,0.8cm)$) -- ++(-0.4cm,-0.3cm);
		\draw[-{Latex[length=2.5mm]}, line width=2pt]
		($(Bz.south west)+(1.4cm,0.8cm)$) -- ++(-0.4cm,-0.3cm);
		\draw[-{Latex[length=2.5mm]}, line width=2pt]
		($(Gz.south west)+(1.4cm,0.8cm)$) -- ++(-0.4cm,-0.3cm);
		\draw[-{Latex[length=2.5mm]}, line width=2pt]
		($(Lz.south west)+(1.4cm,0.8cm)$) -- ++(-0.4cm,-0.3cm);
		
		\draw[-{Latex[length=2.5mm]}, line width=2pt]
		($(Az.south west)+(3.4cm,0.8cm)$) -- ++(-0.0cm,-0.5cm);
		\draw[-{Latex[length=2.5mm]}, line width=2pt]
		($(Bz.south west)+(3.4cm,0.8cm)$) -- ++(-0.0cm,-0.5cm);
		\draw[-{Latex[length=2.5mm]}, line width=2pt]
		($(Gz.south west)+(3.4cm,0.8cm)$) -- ++(-0.0cm,-0.5cm);
		\draw[-{Latex[length=2.5mm]}, line width=2pt]
		($(Lz.south west)+(3.4cm,0.8cm)$) -- ++(-0.0cm,-0.5cm);
		
	\end{tikzpicture}
	\caption{Rendered depth estimations (top), closeups (middle), RGB and Lidar DSM ground truth (bottom). Our geometry regularization technique reduces high-frequency artifacts caused by overfitting to individual training images.}
	\label{fig:intro}
	\vspace{-0.1cm}
\end{figure}
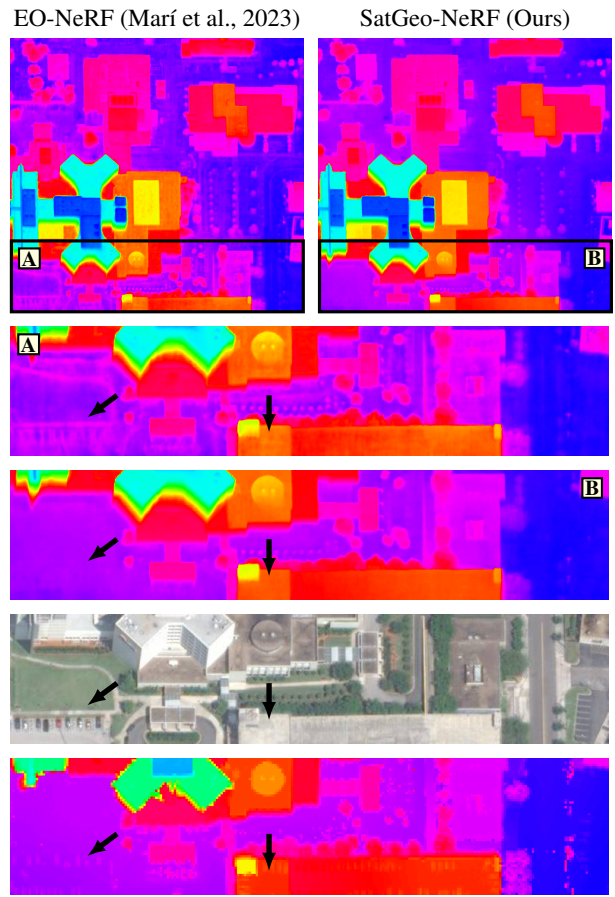

We observe that \emph{NeRFs} tackle cross-view inconsistencies by subtly warping geometry of nominally flat regions, producing wave-like geometric artifacts. Such effects are evident in \cref{fig:intro}; the state-of-the-art \emph{EO-NeRF} depth map shows artificial structures on walkways and flat roof surfaces.
In this work, we therefore focus on geometrical regularization based loss functions to mitigate the artifacts caused by geometrical overfitting.

Urban scenes naturally contain large nominally flat areas such as streets, parking lots or roof structures.
Because these surfaces are expected to be perpendicular to gravity, we propose aligning approximated local surface planes with the axis of gravity to provide a physically grounded prior that suppresses wave-like geometric artifacts.
We apply the regularizer scene-wide and rely on the dominant photometric objective to preserve detail and avoid over-smoothing.

\subsection{Contributions}
\label{subsec:contributions}

\begin{itemize}[
	labelwidth=!,
	labelindent=\labelwidth,
	leftmargin=0pt,
	itemindent=1.2em,
	labelsep=.5em,
	nosep
	]
	\itemsep-0.09em
	\item We propose \modelname{}, a geometrically regularized \emph{NeRF} for satellite  imagery, featuring three model-agnostic regularization techniques to mitigate overfitting-induced geometric artifacts commonly observed in current state-of-the-art models.
	\item The first model-agnostic regularization technique, \emph{Gravity-Aligned Planarity Regularization}, provides a physically grounded prior by aligning depth-inferred, approximated surface normals with the axis of gravity, suppressing wave-like geometric artifacts. 
	This regularization approach connects adjacent rays via a corresponding surface approximation, facilitating gradient flow across the participating rays.
	\item The second model-agnostic regularization technique, \emph{Granularity Regularization} for satellite-domain \emph{NeRFs}, enforces a coarse\hyp{}to\hyp{}fine geometry\hyp{}learning scheme. We also demonstrate the continued benefit of the third model-agnostic regularizer, \emph{Depth\hyp{}Supervised Regularization}, which regularizes depth during the initial training stage using sparse 3D points.
	\item \modelname{} achieves state-of-the-art results on the DFC\hyp{}2019 benchmark scenes, improving the MAE by 14.0\% percent and 11.4\% relative to the previous state-of-the-art.
\end{itemize}

\section{Related Works}
\label{sec:related_works}

The flexibility of the learned, differentiable rendering \emph{NeRFs} provide has shown to be beneficial in handling many of the challenges of multi-date satellite data.
Notable contributions include
\emph{Shadow-NeRF} \cite{snerf} as one of the first adaptations of \emph{NeRF} to the satellite domain, mainly proposing to render shadows as their own lighting component based on the solar position. 
\emph{Sat-NeRF} \cite{satnerf} introduces transient uncertainty and utilizing the satellite-domain-specific \emph{Rational-Polynomial-Camera} (\emph{RPC}) models.
\emph{EO-NeRF} \cite{eonerf} and 
\emph{SUNDIAL} \cite{sundial} both improve upon shadow rendering through simulating solar shading by casting additional rays. 
\emph{Season-NeRF} \cite{seasonnerf} allow rendering images across seasonal appearance changes.

Other works propose expanding the adaptable \emph{NeRF} rendering mechanism to other modalities. 
\citet{pseudopan} combine high resolution panchromatic data with lower resolution color information.  
\emph{Semantic-Sat-NeRF} \cite{semantic_satnerf} integrate semantic information into the model, decreasing rendering artifacts caused by moving objects such as vehicles. 

\emph{EO-GS} \cite{eogs} adapt \emph{Gaussian Splatting} \cite{gaussian_splatting} to the satellite domain by approximating the satellite-domain specific cameras as affine projections and encode the scene using a sparse set of 3D-Gaussians.

Geometric regularization within \emph{NeRFs} has so far been explored predominantly in few-shot scenarios. 
\citet{reg_nerf} enforce depth consistency for image patches from novel viewpoints, and \citet{genericflexreg} extend this idea to pixel-wise depth constraints.
\citet{flipnerf} regularize the geometry along orthogonal rays to remove floating artifacts.
\citet{freenerf} propose a coarse-to-fine strategy by gradually increasing the number of frequencies used in the input encoding during training.
\citet{semantic_manhattan} and \citet{semantic_manhattan_indoor} leverage semantic priors to impose localized planar constraints, whereas \citet{manhattan_clustering} propose semantic-free planar constraints tailored to indoor scenes by clustering explicit surface normals.

\section{Foundations for Satellite-Specific NeRF}

\subsection{General NeRF Principles}

\emph{NeRFs} \cite{nerf} represent a static three\hyp{}dimensional scene as a continuous volumetric function $\mathcal{F}_{NeRF}$ encoded with an \emph{Multi-Layer Perceptron} (\textit{MLP}) network.
\begin{equation}
	\mathcal{F}_{NeRF}: (\vec{x}, \vec{d}) \mapsto (\sigma, \vec{c})
\end{equation}
For a given 3D scene coordinate $\vec{x} = (x, y, z)$ and viewing direction $\vec{d} = (d_x, d_y, d_z)$, the network predicts a color $\vec{c}$ and density $\sigma$. 

To render images, a ray $\vec{r}(t) = \vec{o} + t \cdot \vec{d}$ is created for each image pixel. Here, $\vec{o}$ and $\vec{d}$ denote the origin and direction vectors. Each ray $\vec{r}$ is discretized into $N$ 3D points $\vec{x}_i$ and used as input for  $\mathcal{F}_{NeRF}$. 
The pixel color $\vec{c}(\vec{r})$ is calculated by aggregating the color values $\vec{c}_i$ predicted for each sampled position $\vec{x}_i$ along the ray $\vec{r}$.
\begin{equation}
	\vec{c}(\vec{r}) = \sum_{i=1}^{N} T_i \alpha_i \vec{c}_i
	\label{eq:basic_nerf_rendering}
\end{equation}
The contribution of each predicted color value $\vec{c}_i$ to the overall ray color $\vec{c}(\vec{r})$ is based on its opacity $\alpha_i$ and transmittance $T_i$.
\begin{equation}
	\alpha_i = 1 - \exp(-\sigma_i\delta_i) \text{ and } T_i = \prod_{j=1}^{i-1} (1 - \alpha_j)
	\label{eq:nerf_alpha}
\end{equation}
Both attributes depend on the predicted volume density $\sigma_i$ representing the scene geometry. The opacity $\alpha_i$ hereby defines the visibility of the current sample $\vec{x}_i$ based on its density and the distance $\delta_i = t_i - t_{i-1}$ to the previous sample.
The transmittance $T_i$ is based on previous samples visibility and is used to handle occlusions. 

Analog to aggregating the color values in \cref{eq:basic_nerf_rendering} 
the sample position $t_i$ is accumulated along the ray to determine a depth value $\vec{d}(\vec{r})$ representing the distance of the pixel to the scene.
\begin{equation}
	\vec{d}(\vec{r}) = \sum_{i=1}^{N} T_i \alpha_i t_i
	\label{eq:basic_nerf_depth_rendering}
\end{equation}

\emph{NeRFs} minimize the $L2$-Loss between the rendered ray color $\vec{c}(\vec{r})$ and the ground-truth pixel color $\vec{c}_{GT}(\vec{r})$ for rays $\vec{r}\in\mathcal{R}$ randomly sampled from all training views, thereby enforcing multi-view consistency.
\begin{equation}
	L_{color}(\mathcal{R}) = \sum_{\vec{r} \in \mathcal{R}}|| \vec{c}(\vec{r}) - \vec{c}_{GT}(\vec{r}) ||_2^2
	\label{eq:nerf_color_loss}
\end{equation}

\subsection{Satellite-Domain Adapted NeRF}
\label{subsec:eonerf}

To handle the domain-specific requirements related to lighting, transient objects and camera model accuracy \emph{EO-NeRF} \cite{eonerf} proposes an extended volumetric \emph{NeRF} function:
\begin{equation}
	\mathcal{F}_{EONeRF}: (\vec{x}, \vecs{\omega}, \vec{t}_j) \mapsto (\sigma, \vec{c}_a, \vec{a}, \beta, \tau, \vec{A}_j, \vec{b}_j, \vec{q}_j)
\end{equation}
The inputs are the 3D scene coordinate $\vec{x}$, the sun direction $\vecs{\omega}$ and an image-specific embedding vector $\vec{t}_j$ (where $j$ is the image index).
Whereas the geometrical density $\sigma$ is unchanged, the color prediction is decomposed into multiple components: the albedo color $\vec{c}_a$ and the ambient color $\vec{a}$. Additionally, $\vec{A}_j$ and $\vec{b}_j$ describe an affine color transformation between the predicted albedo color and the current image $j$.
To handle transient, non-stationary objects such as vehicles an uncertainty $\beta$ and transient scalar $\tau$ are predicted using the image-specific embedding $\vec{t}_j$.
Finally, to increase camera accuracy the network predicts a 2D coordinate adjustment component $\vec{q}_j$ for each image $j$. %

\subsection{Satellite-Domain-Specific Ray Generation}
\label{subsec:satellite-domain-specific-camera-model}

Analog to recent works such as \citet{satnerf}, \citet{eonerf}, \citet{sundial} and \citet{semantic_satnerf}
the rays for each pixel are created using the satellite-domain specific \emph{Rational-Polynomial-Camera} (\emph{RPC}) \cite{rpc} model.
As it is typically used for georegistration of satellite images, the \emph{RPC} model is directly extracted from the satellite image metadata.
Each pixel is projected into the scene using the \emph{RPC} model based on two predefined scene height boundaries $[h_{max}, h_{min}]$. These points are used as starting and end point of the ray, respectively. This limits the space covered by rays to closely align with the actual scene content and minimizes the sampling of empty areas. 

Based on \citet{eonerf} the projected points are converted into the \emph{Universal Transverse Mercator} (\emph{UTM}) coordinate system. \emph{UTM} divides the earth into zones, projecting each zone onto a planar surface. This approximates the earths ellipsoid shape locally using a flat ground plane. The scene content is therefore aligned with the ground plane, which is defined by the up-vector, i.e. the unit vector along the z-axis

To increase accuracy, a \emph{Bundle Adjustment} preprocessing step is performed across the \emph{RPC} models following the approach in \citet{bundle_adjustment}. Additionally, a learned camera adjustment $\vec{q}_j$ embedding is used. This parameter defines a shift on the XY-axis learned per-image $j$, allowing the network to adjust camera-accuracy during training.

\section{Geometrical Regularization for Satellite Data}
\label{sec:main}

\begin{figure*}[t]
	\centering
	\includegraphics[width=0.9\textwidth]{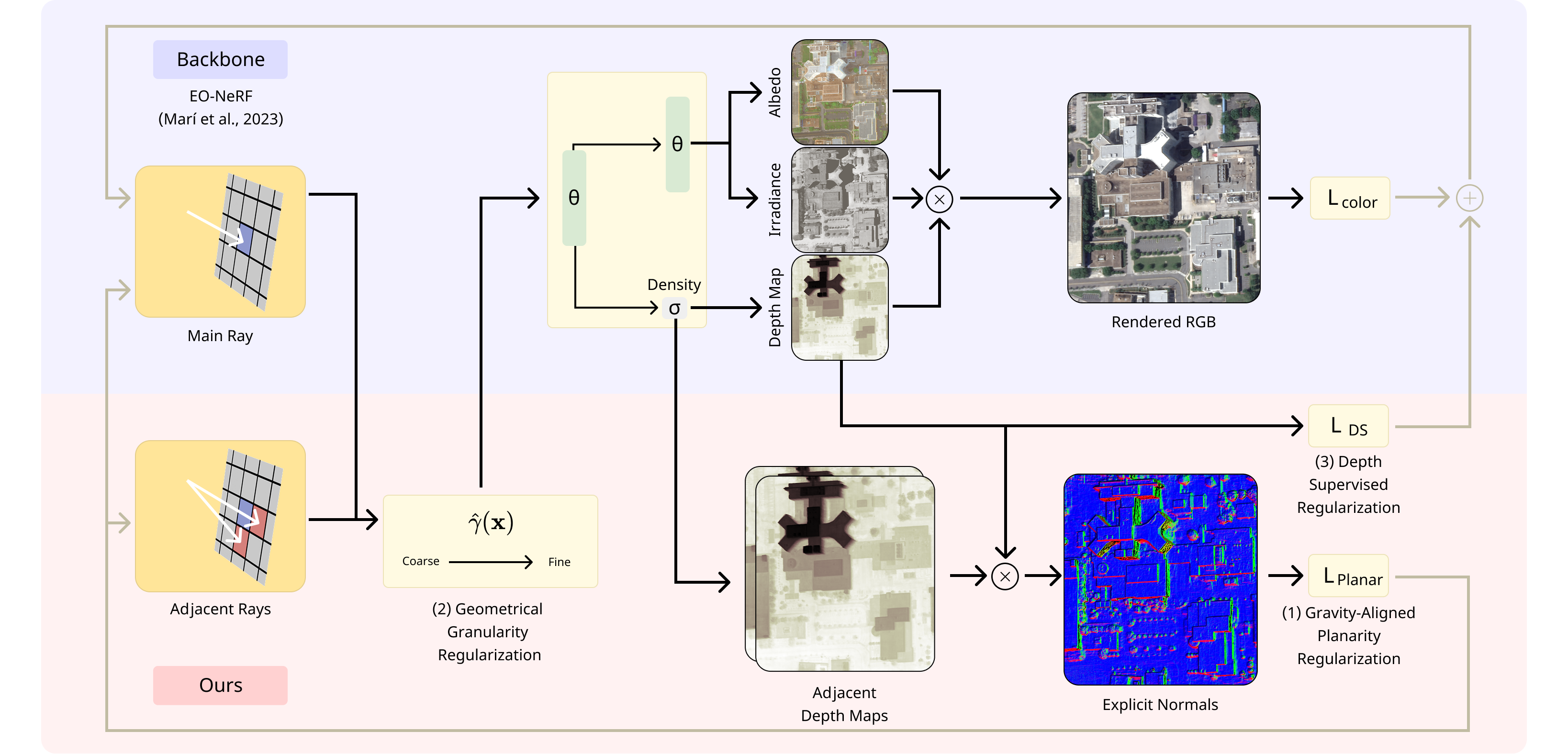}
	\caption{
		We augment a satellite‑domain–adapted backbone (\emph{EO‑NeRF}, \citet{eonerf}) with three model‑agnostic regularizers: 
		(1) \emph{Gravity-Aligned Planarity Regularization}: estimate explicit normals by sampling two adjacent rays per training ray and evaluating depth only; align normals with the axis of gravity to mitigate overfitting and enable cross‑ray smoothing. (2) \emph{Geometrical Granularity Regularization}: coarse‑to‑fine masking of the input frequency encoding.
		(3) \emph{Depth‑Supervised Regularization}: depth constraints from a sparse 3D point cloud obtained from a \emph{Bundle Adjustment} preprocessing step.
	}
	\label{fig:schematic}
\end{figure*}

In this section, we propose \modelname{}, a \emph{NeRF} model for satellite images based on three geometric regularization principles: \emph{Gravity-Aligned Planarity}, \emph{Geometrical Granularity}, and \emph{Depth\hyp{}Supervised}. 
The regularizers are model-agnostic and are transferable to any satellite-domain–adapted \emph{NeRF}. We realize them on \emph{EO}\hyp\emph{NeRF} as described in \cref{subsec:eonerf}, with the interaction shown in \cref{fig:schematic}.

\subsection{Gravity-Aligned Planarity Regularization}

Although photometric supervision alone suffices to train \emph{NeRFs}, it can induce geometry overfitting in single views, producing wave-like artifacts on nominally flat surfaces.
We address this with \emph{Gravity-Aligned Planarity Regularization}, which enforces planarity on locally estimated surface planes using the axis of gravity as a prior.
The term counterbalances the photometric loss by penalizing deviations from local planarity.
Local surface orientation is estimated by computing normals from \emph{NeRF}-predicted depths along adjacent rays. 
We apply the regularizer scene-wide, but optimization is dominated by the photometric term, preserving detail and preventing over-smoothing.

\subsubsection{Explicit Surface Normals}

For each ray $\vec{r}$ we determine two random adjacent rays $\vec{r}_{n_1}, \vec{r}_{n_2}$ from the four immediate neighboring pixels. 
To prevent unstable normal calculation, we make sure that we sample one neighbor from each of the two adjacent pixels in X- and Y-axis. 
All three rays are evaluated to determine their estimated depth values $d(\vec{r})$ as seen in \cref{eq:basic_nerf_depth_rendering}.
The depth is converted into the three-dimensional surface point $\vec{p}_s(\vec{r})$ by following each ray along its direction: $\vec{p}_s(\vec{r}) = \vec{r}(d(\vec{r})) = \vec{o} + d(\vec{r})\vec{d}$.
We approximate a local surface plane by determining the surface normal $\hat{\vec{n}}(\vec{r},\vec{r}_{n_1}, \vec{r}_{n_2})$ based on the three surface points. 
\begin{equation}
	\hat{\vec{n}}(\vec{r},\vec{r}_{n_1}, \vec{r}_{n_2}) = (\vec{p}_s(\vec{r}_{n_1}) - \vec{p}_s(\vec{r})) \times (\vec{p}_s(\vec{r}_{n_2}) - \vec{p}_s(\vec{r}))
\end{equation}
To ensure correct length, we additionally normalize each calculated surface normal, denoted as $\vec n(\vec r,\vec r_{n_1},\vec r_{n_2})$.
To summarize, $\vec{n}$ describes a unit vector orthogonal to the plane defined by the three surface points using the set of rays $\vec{r},$ $\vec{r}_{n_1}$ and $\vec{r}_{n_2}$.
With this process, we approximate local surface planes directly from the learned geometry, closely following the foundational \emph{NeRF} rendering equations in \cref{eq:nerf_alpha,eq:basic_nerf_depth_rendering}.

\begin{figure}[!t]
	\includegraphics[width=1\columnwidth]{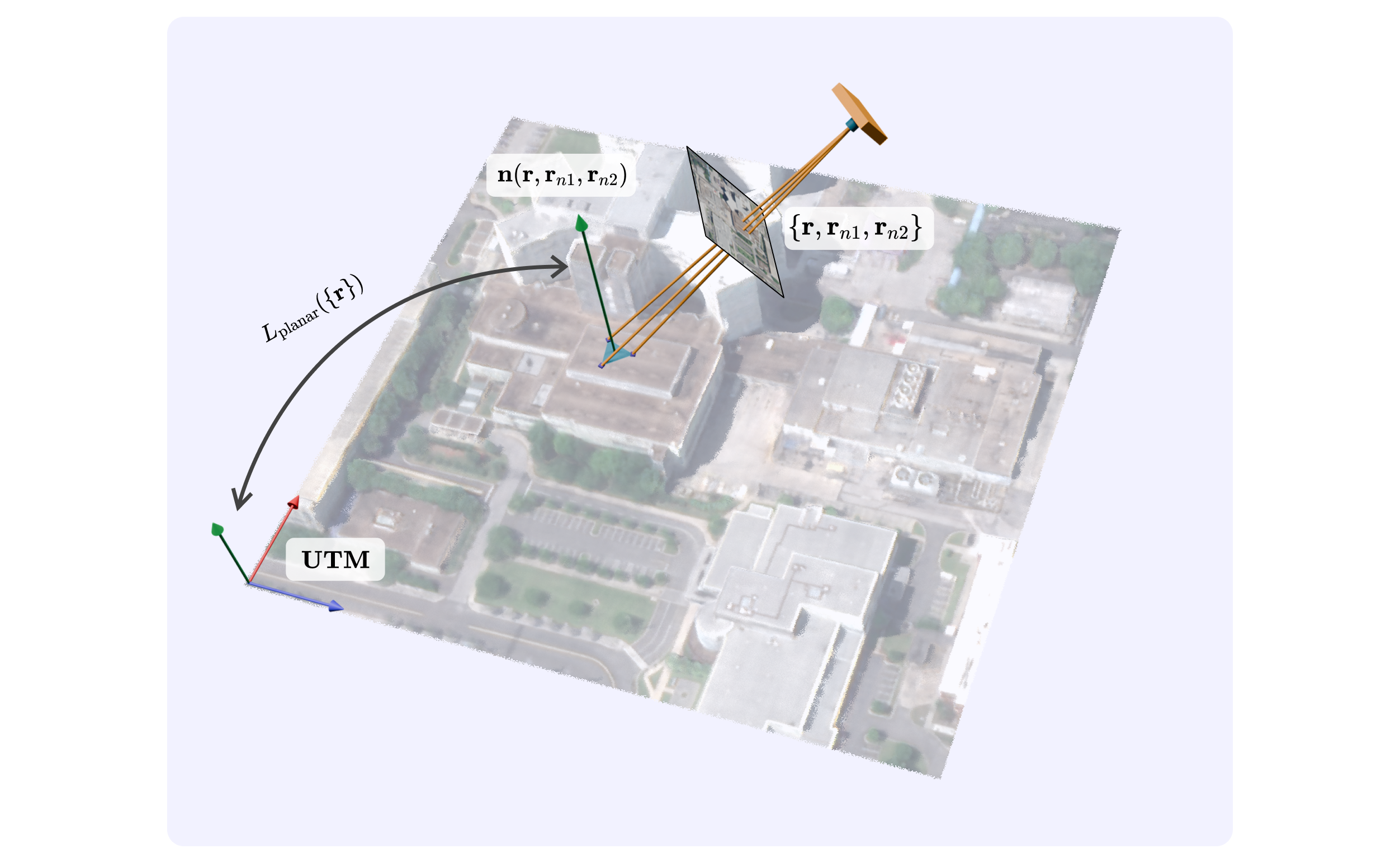}
	\caption{\emph{Gravity-Aligned Planarity Regularization}: A local plane is estimated from three surface points on adjacent rays; its normal is constrained to align with the axis of gravity (approximated as \emph{UTM}-up). }
	\label{fig:local_planarity_reg}
\end{figure}

\subsubsection{Gravity Alignment}
\label{subsec:planar_regularization}

We align the calculated explicit surface normals along the approximated axis of gravity $\vec{g}$ with the regularization loss term $L_{\mathrm{planar}}$.
\begin{equation}
	L_{\mathrm{planar}}(\mathcal{R}) = \sum_{\vec{r} \in \mathcal{R}} ||\vec{g} - \vec{n}(\vec{r}, \vec{r}_{n_1}, \vec{r}_{n_2})||_2
\end{equation}
To estimate the gravity axis $\vec{g}$, we leverage the known alignment of scene content in the \emph{UTM} coordinate system. \emph{UTM} places the ground in the XY-plane, so the Z-axis approximates the inverse gravity vector $\vec{g} = (0,0,1)^T$.

By explicitly computing surface normals with two auxiliary adjacent rays $\vec{r}_{n_1}, \vec{r}_{n_2}$, the \emph{Gravity-Aligned Planarity} loss $L_{planar}$ backpropagates through all samples along the three rays.
Unlike standard \emph{NeRF} training that treats rays independently, this design couples adjacent rays and imposes geometric regularization over an area.

\subsubsection{Minimizing Computational Cost}
\label{subsubsec:computational_cost}

Calculating explicit normals by inferencing two additional rays for each ray $\vec{r}$ increases computational cost by a factor of three if computed naively.
We propose additional measures to reduce the computational cost required by the adjacent rays. 
A major reduction in computational cost is achieved by using only the partial network $\mathcal{F}_{Depth}$, predicting only the geometrical density $\sigma$ based on the 3D coordinate input $\vec{x}$. Output heads related to color, uncertainty or transient prediction are skipped.
\begin{equation}
	\mathcal{F}_{Depth}: (\vec{x}) \mapsto (\sigma)
\end{equation}
We calculate the explicit normals based solely on the depth values, making any outputs related to color unnecessary. We therefore also do not inference the additional shadow rays $\vec{r}_{\mathrm{sun}}$ required for the shadow component $\vec{s}(\vec{r})$.
Additionally, we decrease the number of samples and reduce the length of the ray to minimize sampling empty space. 
Similar to the normal calculation, we assume that the selected points define a local (flat) approximation of the surface. We therefore derive the length of the adjacent rays to cover the immediate range surrounding the predicted depth $\vec{d}(\vec{r})$ of the main ray $\vec{r}$.

With the main ray $\vec{r}$ covering the length $t_l = t_{far} - t_{near}$, we define the partial ray length as $t_n = t_l \cdot p_n$ with $p_n \in [0, 1]$ as scaling factor.
The partial adjacent rays $\vec{r}_n$ are then defined by shifting the near and far points based on the partial ray length $t_n$.
\begin{equation}
	\vec{r}_{n}(t) = \vec{o}_n + t\vec{d} \text{ with } t \in [t_{near} + \tfrac{t_l - t_n}{2}, t_{far} - \tfrac{t_l - t_n}{2}]
\end{equation}
To keep a similar spacing of the samples along the rays, the number of samples $N_n$ is reduced to $N_n = N \cdot p_n$.

\subsection{Geometrical Granularity Regularization}

The general concept of \emph{NeRFs} \cite{nerf} struggles to capture high-frequency color and geometric detail when operating directly on low-frequency sample coordinates $\vec{x}$. To address this, \citet{nerf} apply a high-frequency positional encoding $\gamma(\vec{x})$ composed of sine and cosine functions at $L$ increasing frequencies.
\begin{equation}
	\begin{aligned}
		\vecs{\gamma}(\vec{x}) 
		&= (\sin(2^0 \pi \vec{x}), \cos(2^0 \pi \vec{x}), \dots, \\
		&\quad \sin(2^{L-1}\pi \vec{x}), \cos(2^{L-1}\pi \vec{x}))
	\end{aligned}
\end{equation}
However, we observe that jointly learning coarse and fine scales induces geometric artifacts in satellite imagery
We therefore introduce a \emph{Geometrical Granularity Regularization} that masks encoding frequencies during training, based on \citet{freenerf}, thereby enforcing a coarse-to-fine schedule at negligible extra computational cost.
Based on the current training iteration $t$, final regularization iteration $T$ and maximum encoding frequency $L$, a bitmask $\vecs{\alpha}(t, T, L)$ is calculated. We apply the bitmask $\vecs{\alpha}$ to the input encoding $\vecs{\gamma}$ to form the regularized input encoding $\vecs{\gamma}'$:
\begin{equation}
	\vecs{\alpha}_i(t,T,L) = 
	\begin{cases}
		1, & i <  2(\frac{t \cdot L}{T} + b),\\
		0, & i \ge 2(\frac{t \cdot L}{T} + b).
	\end{cases}
\end{equation}
\begin{equation}
	\vecs{\gamma}'(\vec{x}, t) = \vecs{\gamma}(\vec{x})~\odot~\vecs{\alpha}(t, T, L)
\end{equation}
In practice, this mask increases the visibility by one $sine$ or $cosine$ frequency each time training advances by a fixed amount, defined by the final regularization iteration $T$.
The bias term $b$ specifies the initial number of active frequency bands, providing sufficient low-frequency signals at the start.

Originally introduced by \citet{freenerf} for few-shot scenarios, we employ a short regularization window $T$ with a high initial bias $b$ as a geometrical granularity regularization, reducing geometric error in satellite scenes with negligible overhead. 
Because the schedule is brief, we substitute gradual unmasking of input frequencies with a binary on/off threshold.

\subsection{Depth Supervised Regularization}

To increase accuracy of the used camera model, \emph{Bundle Adjustment} is performed across all \emph{RPC}-cameras \cite{bundle_adjustment} of a given scene by minimizing multi‑view reprojection error over image-based correspondences.
As byproduct a set of sparse points is derived from image features. 
In contrast to recent work, we observe that guiding the network during the initial stage of training with this known sparse point cloud positively impacts performance.

We construct an additional set of rays $\mathcal{R}_{\mathrm{DS}}$ for the derived 3D points and its corresponding image pixel.
The \emph{Depth Supervised Regularization} loss $L_{\mathrm{DS}}(\mathcal{R}_{\mathrm{DS}})$ compares the predicted depth $d(\vec{r})$ with the depth based on the known 3D position $\vec{X}(\vec{r})$ and ray origin $\vec{o}(\vec{r})$, weighted with the reprojection error $w(\vec{r})$.
\begin{equation}
	L_{\mathrm{DS}}(\mathcal{R}_{\mathrm{DS}}) = \sum_{\vec{r} \in \mathcal{R}_{\mathrm{DS}}} w(\vec{r}) \left( d(\vec{r}) - \left\| \vec{X}(\vec{r}) - \vec{o}(\vec{r}) \right\|_2 \right)^2
\end{equation}
Originally proposed by \citet{satnerf}, expanding works such as \citet{eonerf} and \citet{sundial} chose to forfeit this additional regularization, arguing that image-based loss terms are sufficient. However, similar to the \emph{Gravity-Aligned Planarity Regularization} described in \cref{subsec:planar_regularization}, we instead argue that any known prior should be utilized to guide the network and increase the quality of the learned geometry. The positive effects of \emph{Depth Supervised Regularization} are shown in \cref{tab:eval_ablation_study}.

\section{Implementation}

We use \emph{EO-NeRF} \cite{eonerf} as backbone, expanding the architecture with three additional regularization terms. 
At the time of our experiments, the official implementation of \emph{EO-NeRF} was not publicly available, and we therefore reimplement the method. As \citet{eonerf} describe only the high-level network structure, we base our concrete architecture on the publicly available model of \citet{satnerf}.
We split the network into a main feature backbone and several specialized heads. The backbone additionally outputs the density $\sigma$, while the heads are small one-layer MLPs fed by an additional projection layer. 
The backbone consists of $8$ layers with a width of $256$ and heads use one layer with a width of $128$. We employ the ReLU activation function between hidden layers, sigmoid for all color outputs, and softplus for $\beta$ and $\sigma$ prediction.

Following established satellite specific training strategies (\citet{satnerf}, \citet{eonerf}, \citet{sundial} and \citet{semantic_satnerf}), we train with the standard \emph{NeRF} color loss from \cref{eq:nerf_color_loss} for the first two epochs and afterwards switch to the uncertainty-aware loss from \citet{nerf-in-the-w}.
The \emph{Gravity-Aligned Planar Regularization} $L_{\mathrm{planar}}$ is turned on at epoch three once a coarse scene representation is learned, and its weight is empirically chosen as $\lambda_{\mathrm{planar}} = 0.1$.
The \emph{Depth Supervised Regularization} $L_{\mathrm{DS}}$ is enabled for the first $25$\% of iterations, guiding the network in the initial learning stage with a weight of $\lambda_{\mathrm{DS}} = 1000$.

We set the positional and directional encoding frequency ranges to $L_p = 10$ and $L_d = 4$, following \citet{nerf}. We apply \emph{Geometrical Granularity Regularization} to the positional encoding frequencies, using a starting bias $b = L_p/2$, which enables half of the available frequencies initially. We use a short regularization window $T = \lfloor 0.10 \cdot \textit{total\_iterations} \rfloor$, as proposed by \citet{freenerf} for non–few-shot settings. Each ray is sampled with $N = 128$ samples.

All experiments are optimized over $300.000$ iterations using an Adam optimizer with an initial learning rate of $5e-4$ and a batch size of $1024$.
A single training takes approximately $10$ hours using an NVIDIA RTX 4090.

\section{Experiments}
\label{sec:eval}

In this section, we evaluate the impact of our proposed \modelname{} on reconstruction quality with respect to both learned geometry and image reproduction. Additionally, we provide an ablation study covering all proposed regularization terms. Finally, 
we demonstrate the reduction of computational costs for explicit surface estimation by evaluating only partial segments of adjacent rays.

\subsection{Dataset}

Analog to previous works we evaluate our model on a widely used subset of four scenes from the 2019 IEEE GRSS Data Fusion Contest (DFC2019) \cite{dfc2019}, comprising multi-date imagery (2014–2016) over Jacksonville, Florida. Each site includes 10–20 captures at $\SI{0.3}{\metre}$ ground sampling distance with varying off‑nadir angles. The imagery is pan\hyp{}sharpened from multispectral and panchromatic sources and preprocessed to 8‑bit true‑color RGB. Experiments are conducted on a $\SI[parse-numbers=false]{256\times256}{\square\metre}$ area aligned to the provided LiDAR Digital Surface Model (DSM). We perform an initial bundle adjustment of the \emph{RPC} models of each scene; the optimized RPCs are used for all methods, including baselines.

\subsection{Comparison with Baseline}

We conduct a comparative evaluation against state-of-the-art models such as \emph{EO-NeRF} \cite{eonerf} and \emph{EO-GS} \cite{eogs}. 
At the time of experiments, the official \emph{EO-NeRF} implementation was not publicly available. We therefore rely on our reimplementation (\emph{EO-NeRF*}).
Our reproduced results do not reach the performance reported by \citet{eonerf}. 
This discrepancy in performance is in line with other reimplementations of \emph{EO-NeRF} such as \citet{sundial}. 

\begin{figure}
	\centering
	\subfigure[\emph{EO-NeRF*: 29.20}]{\includegraphics[width=.49\linewidth]{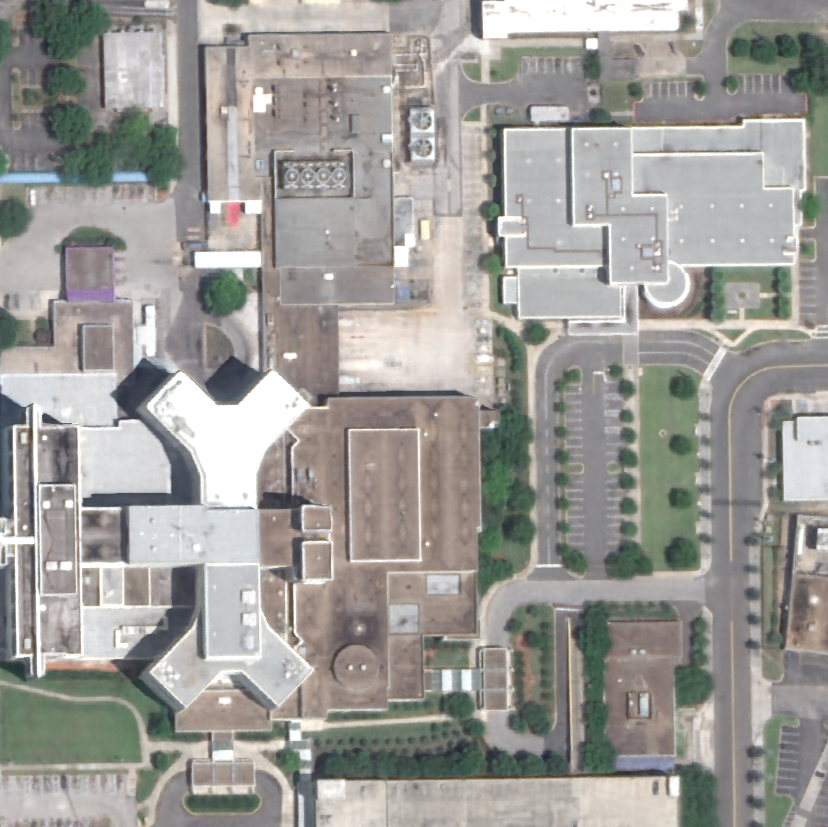}}
	\hfill
	\subfigure[\modelname{} (Ours): 27.26]{\includegraphics[width=.49\linewidth]{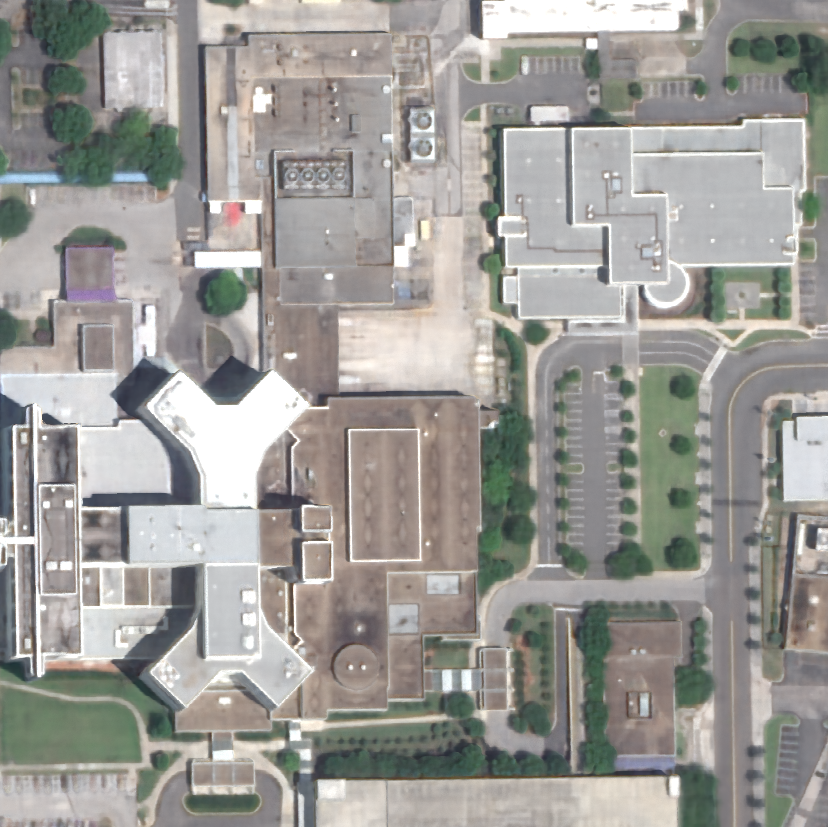}}
	\hfill
	\caption{
		Rendered RGB images with their respective PSNR values. To human observers, the $6.6$\% relative PSNR difference is not perceptually noticeable under normal viewing.}
	\label{fig:results_rgb}
\end{figure}

\begin{table*}
	\begin{center}
		{
			\footnotesize\tabcolsep3pt
			\begin{tabular}{c|p{3cm}|llll|l|llll|l}
				\toprule
				Foliage & & \multicolumn{5}{c|}{\textbf{MAE [m] $\downarrow$}} &\multicolumn{5}{c}{\textbf{PSNR $\uparrow$}} \\
				Mask & \multicolumn{1}{l|}{Model} & \multicolumn{1}{c}{004} & \multicolumn{1}{c}{068} & \multicolumn{1}{c}{214} & \multicolumn{1}{c}{260} & \multicolumn{1}{|c|}{Mean} & \multicolumn{1}{c}{004} & \multicolumn{1}{c}{068} & \multicolumn{1}{c}{214} & \multicolumn{1}{c}{260} & \multicolumn{1}{|c}{Mean}\\
				
				\midrule
				
				\multirow{5}{*}{\crossmark} & SatNeRF \textit{\tiny \cite{satnerf}} & \secondplace{1.39} \silvermedal & 1.45 & 2.57 & 1.77 & 1.80 & 32.11 & 29.60 & 28.22 & 29.20 & 29.78 \\
				
				& EO-NeRF* \textit{\tiny \cite{eonerf}} & 1.57 & \secondplace{1.23} \silvermedal & 2.43 & \secondplace{1.63} \silvermedal & 1.72 & 29.20 & 27.33 & 26.51 & 27.28 & 27.58 \\
				& EO-GS \textit{\tiny \cite{eogs}}& 1.45 & 1.25 & \secondplace{2.30} \silvermedal & 1.69 & \secondplace{1.67} \silvermedal & 33.55 & 28.71 & 26.16 & 28.78 & 29.30 \\
				
				\cmidrule{2-12}

				& \model{} (Ours) & \firstplace{1.30} \goldmedal & \firstplace{1.08} \goldmedal & \firstplace{2.13} \goldmedal & \firstplace{1.39} \goldmedal & \firstplace{1.48} \goldmedal & 27.26 & 26.68 & 26.02 & 26.02 & 26.50 \\
				& {\tiny Compared to EO-NeRF*} & \improvement{17.2} & \improvement{12.2} & \improvement{12.4} & \improvement{14.7} & \improvement{14.0} & \decrease{6.6} & \decrease{2.4}  & \decrease{1.8}  & \decrease{4.6} & \decrease{3.9} \\

				\midrule

				\multirow{5}{*}{\checkmark} & SatNeRF \textit{\tiny \cite{satnerf}} & 1.02 & 1.43 & 2.62 & 1.73 & 1.70 & 31.76 & 29.42 & 28.12 & 28.94 & 29.56 \\
				& EO-NeRF* \textit{\tiny \cite{eonerf}} &  \secondplace{0.91} \silvermedal & \secondplace{1.21} \silvermedal & 2.42 & \secondplace{1.40} \silvermedal & \secondplace{1.49} \silvermedal & 29.60 & 27.08 & 26.37 & 27.12 & 27.54\\
				& EO-GS \textit{\tiny \cite{eogs}} & 0.92 & 1.22 & \secondplace{2.25} \silvermedal & 1.55 & \secondplace{1.49} \silvermedal & 35.88 & 28.92 & 26.35 & 29.32 & 30.12 \\
				
				\cmidrule{2-12}

				& \model{} (Ours)  & \firstplace{0.78} \goldmedal & \firstplace{1.05} \goldmedal & \firstplace{2.10} \goldmedal & \firstplace{1.16} \goldmedal  & \firstplace{1.27} \goldmedal & 27.28 & 26.42 & 25.88 & 25.70 & 26.32 \\
				& {\tiny Compared to EO-NeRF*} & \improvement{14.3} & \improvement{13.2} & \improvement{13.2} & \improvement{17.1} & \improvement{14.8} & \decrease{7.8} & \decrease{2.4}  & \decrease{1.9}  & \decrease{5.2} & \decrease{4.4} \\
				
				\bottomrule
			\end{tabular}
		}
	\end{center}
	\caption{Evaluation of our proposed \modelname{} with state-of-the-art baselines. Our proposed regularization are able to improve the MAE by a mean of 14.0\% on the DFC2019 dataset \cite{dfc2019} with only minimal impact on image render quality. 
}
	\label{tab:eval_main}
\end{table*}

\begin{table*}[t]
	\begin{center}
		{
			\footnotesize\tabcolsep3pt
			\begin{tabular}{ccc|llll|l|llll|l}
				\toprule
				\multicolumn{3}{c|}{\textbf{Regularization}} & \multicolumn{5}{c|}{\textbf{MAE [m] $\downarrow$}} &\multicolumn{5}{c}{\textbf{PSNR $\uparrow$}} \\
				Granularity & Planarity & Depth & \multicolumn{1}{c}{004} & \multicolumn{1}{c}{068} & \multicolumn{1}{c}{214} & \multicolumn{1}{c}{260} & \multicolumn{1}{|c|}{Mean} & \multicolumn{1}{c}{004} & \multicolumn{1}{c}{068} & \multicolumn{1}{c}{214} & \multicolumn{1}{c}{260} & \multicolumn{1}{|c}{Mean}\\
				
				\midrule
				
				\crossmark & \crossmark & \crossmark &  1.57 & 1.23 & 2.43 & 1.63 & 1.72 & 29.20 & 27.33 & 26.51 & 27.28 & 27.58\\
				
				\checkmark & \crossmark & \crossmark & 1.39 & 1.16  & 2.20 & \secondplace{1.44} \silvermedal & \secondplace{1.55} \silvermedal & 28.60 & 26.98 & 26.22 
				& 26.54 & 27.09 \\
				
				\checkmark & \checkmark & \crossmark & 1.72 & \firstplace{1.06} \goldmedal & \firstplace{2.10} \goldmedal & 1.56 & 1.61 & 28.72	  & 27.18   & 26.37   & 26.93  & 27.30 \\
				\checkmark & \checkmark & \checkmark & \firstplace{1.30} \goldmedal & \secondplace{1.08} \silvermedal & \secondplace{2.13} \silvermedal & \firstplace{1.39} \goldmedal & \firstplace{1.48} \goldmedal & 27.26 & 26.68 & 26.02 & 26.02 & 26.50 \\
				
				\bottomrule
				
			\end{tabular}
		}
	\end{center}
	\caption{Ablation study demonstrating improvements in reconstruction quality from our proposed regularization terms.
		}
	\label{tab:eval_ablation_study}
\end{table*}

To assess the quality of the reconstructed geometry, we report the \emph{Mean Altitude Error} (MAE). 
For each training image, we convert predicted depth to a \emph{Digital Surface Model} (DSM), align the model to a ground‑truth LiDAR DSM, and compute the per‑view altitude error. The final scene score represents to the mean across all training views.
In combination with the wide range of off-nadir angles in the training data this yields a robust geometric evaluation.
While \emph{EO-GS} originally reports an altitude error for a single nadir view, we extend their evaluation to all training views for a fair comparison.
Because PSNR is unreliable for novel views due to ambient illumination changes and transient effects, we report PSNR only on the input (training) views.

The quantitative results are presented in \cref{tab:eval_main}. Across all four scenes our proposed regularization are able to decrease the MAE by a mean of 14.0\% compared to the state-of-the-art \emph{EO-NeRF} \cite{eonerf} method. As the network is not able to overoptimize the geometry as freely, the PSNR value for the training views drops slightly by 3.9\%. As shown in \cref{fig:results_rgb}, this difference is not perceptually noticeable to human observers.

Whereas urban scene content remains relatively static across images, vegetation such as trees feature increased variance due to seasonal changes. We apply a foliage mask during evaluation to filter out all vegetation to evaluate the impact of our proposed regularization on urban scene content specifically.
The semantic masks for the Lidar Ground-Truth are provided by \emph{EO-GS} \cite{eogs} and for the RGB images by \emph{Semantic-SatNeRF} \cite{semantic_satnerf}. 
We are able to decrease the MAE by 14.8\% for urban scene content compared to \emph{EO-NeRF} \cite{eonerf}.

\begin{figure*}[h]
	\centering
	\subfigure[W/o Planarity Regularization]{\includegraphics[width=.24\linewidth]{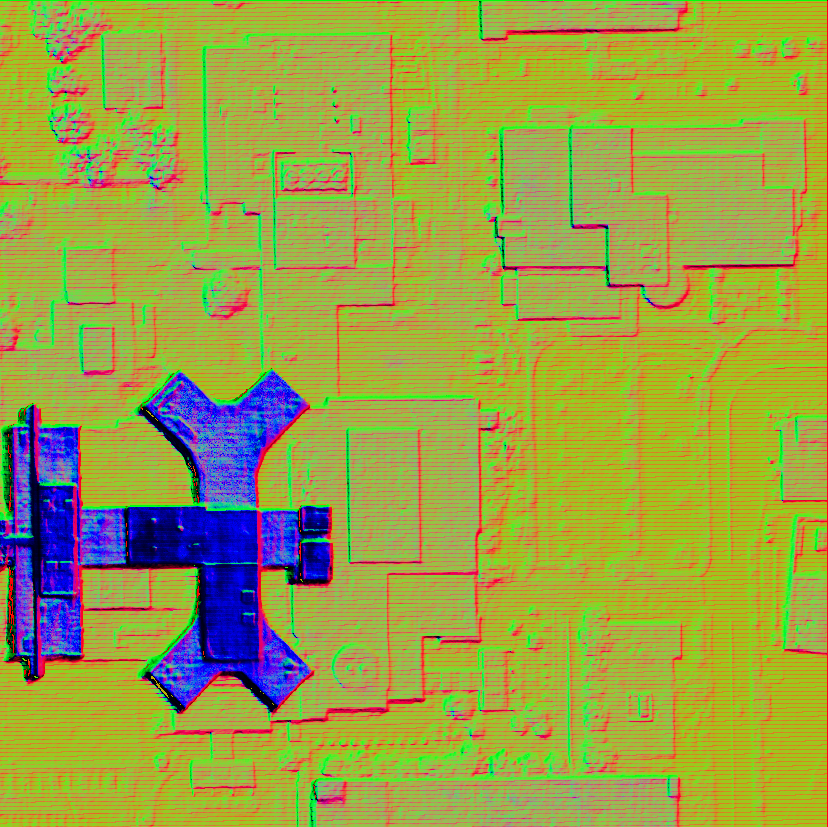}}
	\hfill
	\subfigure[With Planarity Regularization]{\includegraphics[width=.24\linewidth]{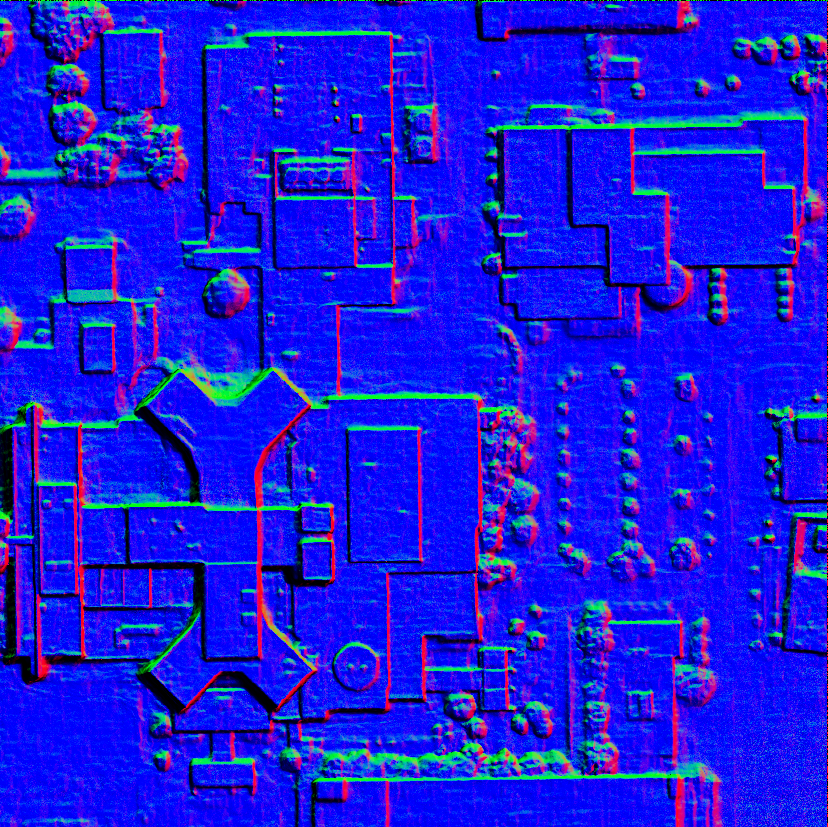}}
	\hfill
	\subfigure[W/o Planarity Regularization]{\includegraphics[width=.231\linewidth]{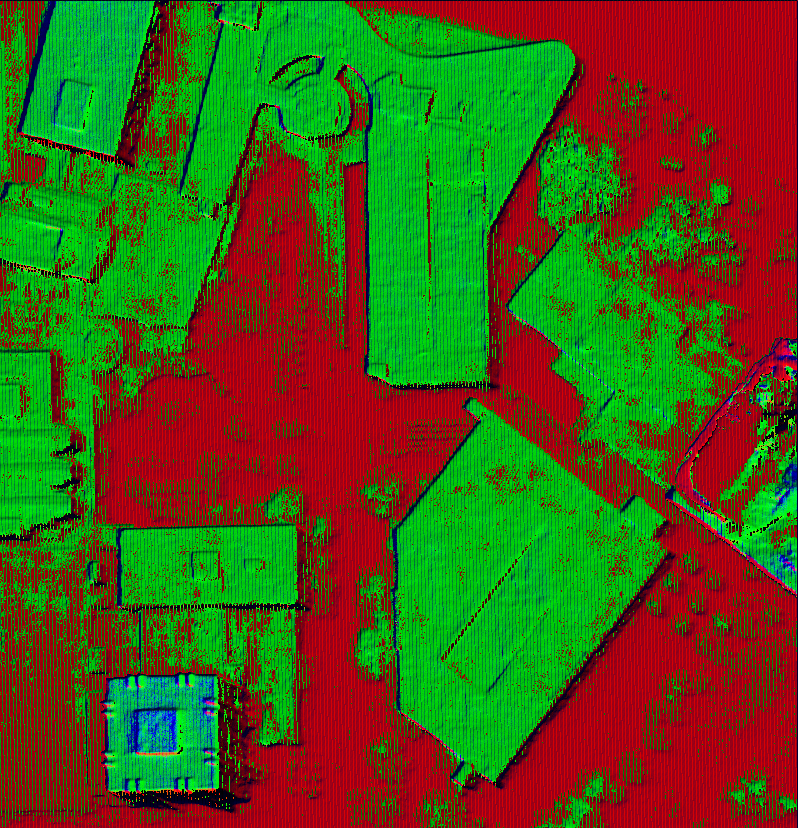}}
	\hfill
	\subfigure[With Planarity Regularization]{\includegraphics[width=.231\linewidth]{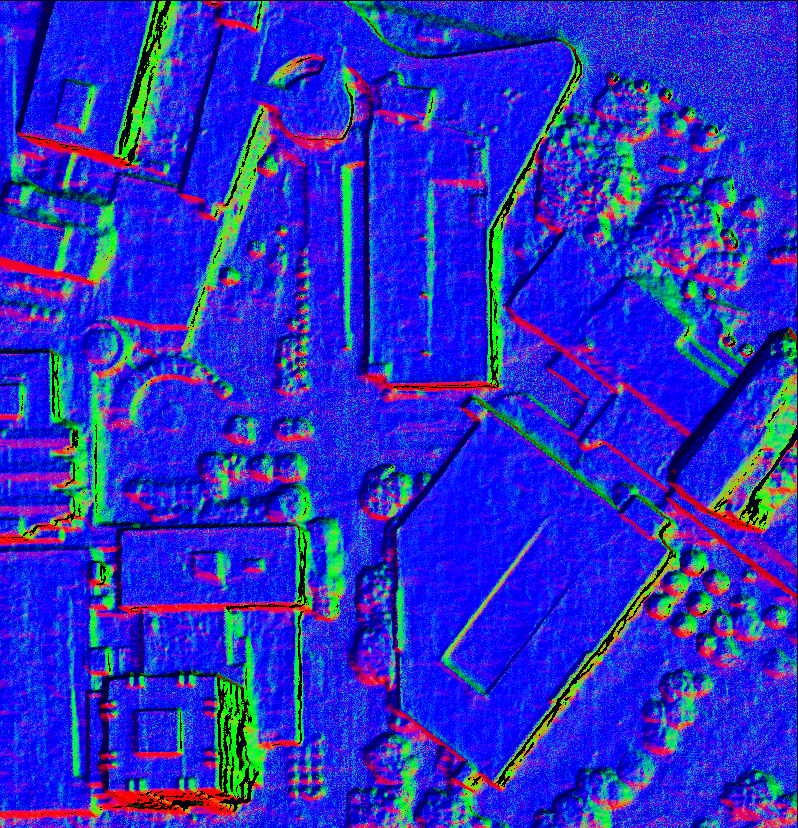}}
	\hfill
	\caption{Explicit surface normal vectors determined based on predicted depth values of three adjacent rays; colors show normalized orientation (e.g., blue = upright). \emph{Planarity Regularization} increases quality through alignment with the axis of gravity. 
}
	\label{fig:results_neighbours}
\end{figure*}

\begin{figure*}[h]
	\newcommand{\imgwidth}{0.23\linewidth}
	\centering
	
	\begin{minipage}{0.015\linewidth}
		\centering
		\rotatebox[origin=center]{90}{}
	\end{minipage}
	\begin{minipage}{0.98\linewidth}
		\centering
		\begin{minipage}{\imgwidth}\centering JAX\_004\end{minipage}
		\hfill
		\begin{minipage}{\imgwidth}\centering JAX\_068\end{minipage}
		\hfill
		\begin{minipage}{\imgwidth}\centering JAX\_214\end{minipage}
		\hfill
		\begin{minipage}{\imgwidth}\centering JAX\_260\end{minipage}
	\end{minipage}
	
	\begin{minipage}{0.015\linewidth}
		\centering
		\rotatebox[origin=center]{90}{Lidar Ground Truth}
	\end{minipage}
	\begin{minipage}{0.98\linewidth}
		\centering
		\subfigure{\includegraphics[width=\imgwidth]{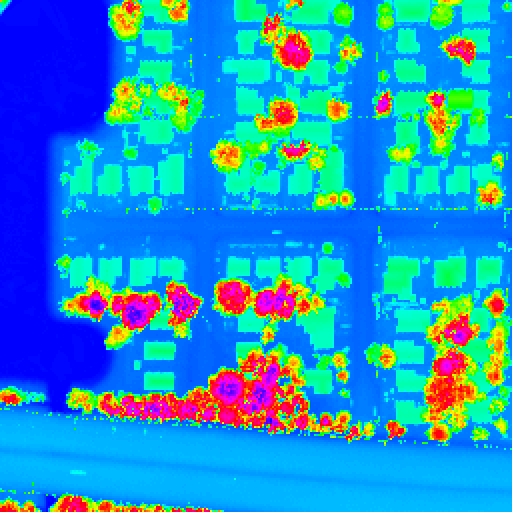}}
		\hfill
		\subfigure{\includegraphics[width=\imgwidth]{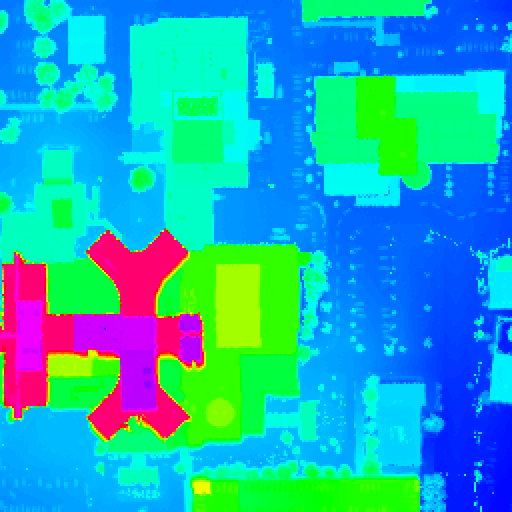}}
		\hfill
		\subfigure{\includegraphics[width=\imgwidth]{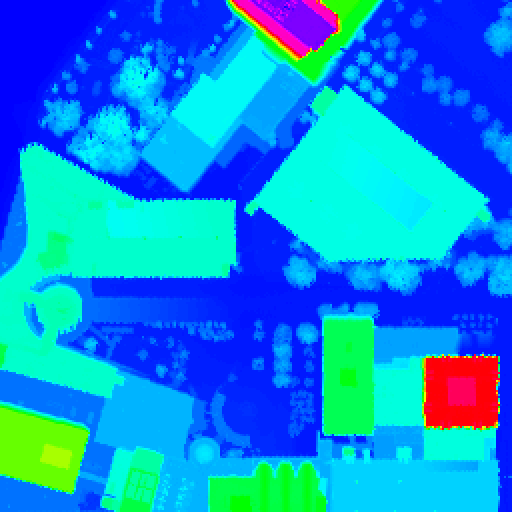}}
		\hfill
		\subfigure{\includegraphics[width=\imgwidth]{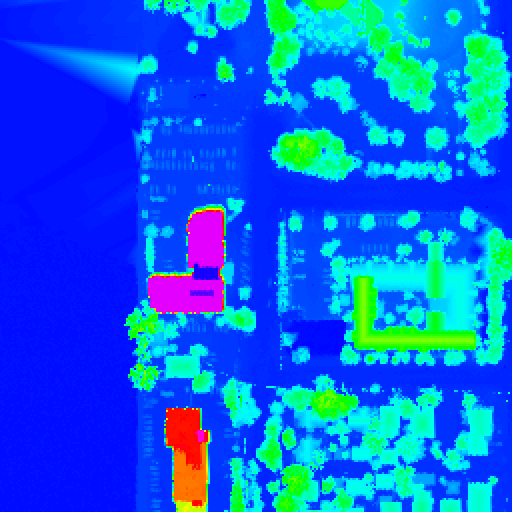}}
	\end{minipage}
	
	\begin{minipage}{0.015\linewidth}
		\centering
		\rotatebox[origin=center]{90}{EO-NeRF*}
	\end{minipage}
	\begin{minipage}{0.98\linewidth}
		\centering
		\subfigure{\includegraphics[width=\imgwidth]{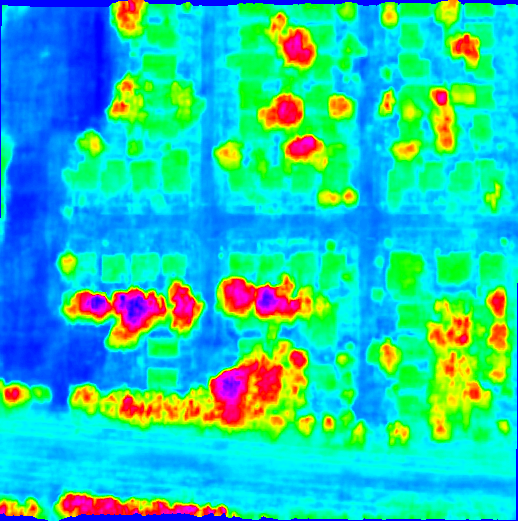}}
		\hfill
		\subfigure{\includegraphics[width=\imgwidth]{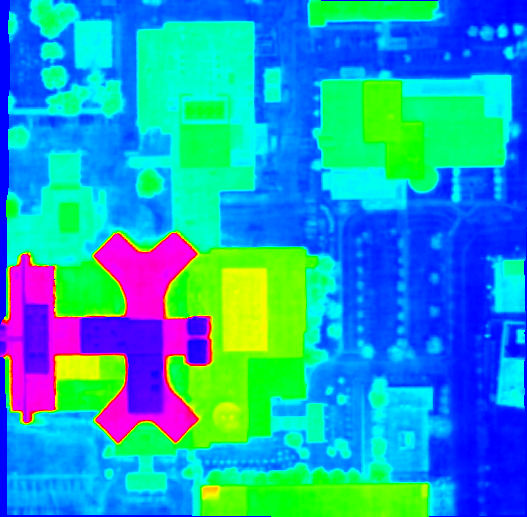}}
		\hfill
		\subfigure{\includegraphics[width=\imgwidth]{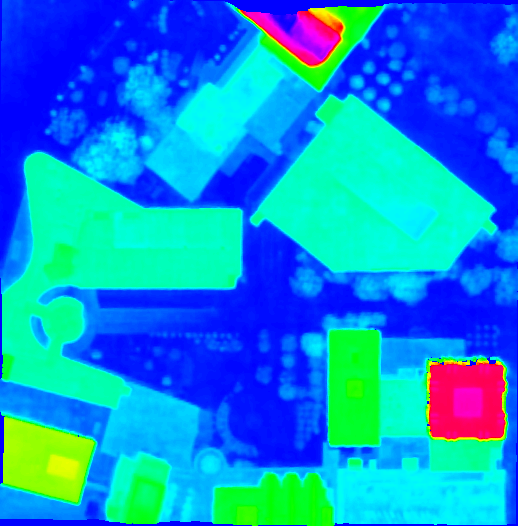}}
		\hfill
		\subfigure{\includegraphics[width=\imgwidth]{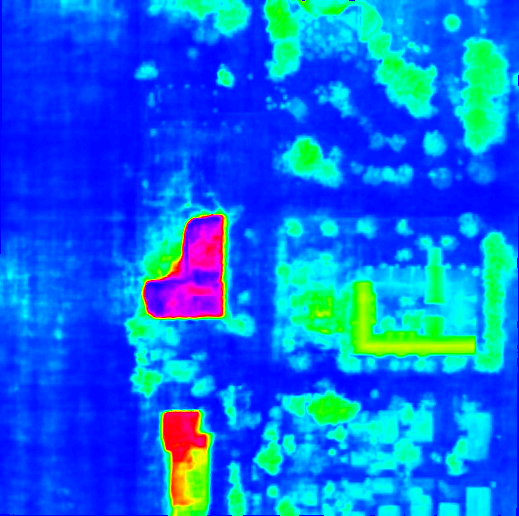}}
	\end{minipage}
	
	\begin{minipage}{0.015\linewidth}
		\centering
		\rotatebox[origin=center]{90}{EO-GS}
	\end{minipage}
	\begin{minipage}{0.98\linewidth}
		\centering
		\subfigure{\includegraphics[width=\imgwidth]{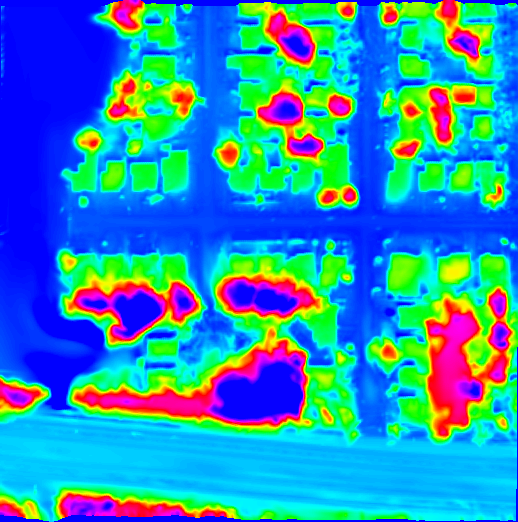}}
		\hfill
		\subfigure{\includegraphics[width=\imgwidth]{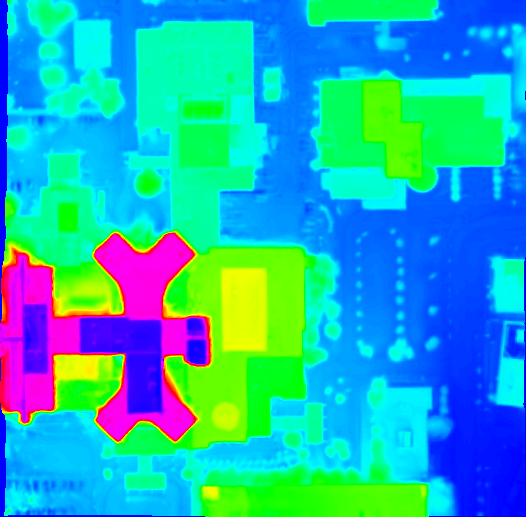}}
		\hfill
		\subfigure{\includegraphics[width=\imgwidth]{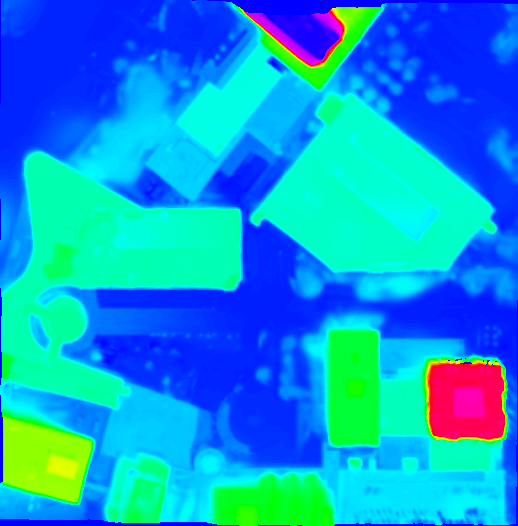}}
		\hfill
		\subfigure{\includegraphics[width=\imgwidth]{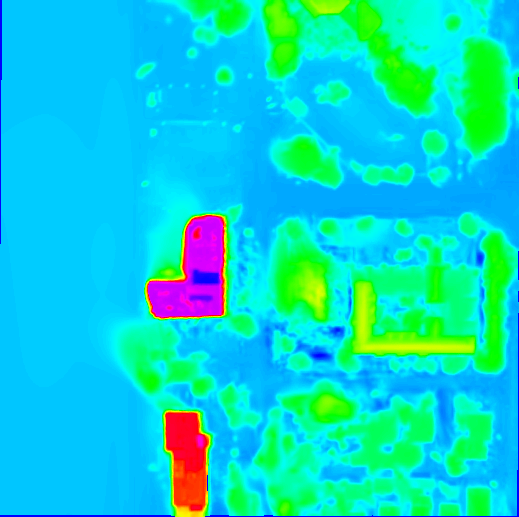}}
	\end{minipage}
	
	\begin{minipage}{0.015\linewidth}
		\centering
		\rotatebox[origin=center]{90}{\model{} (Ours)}
	\end{minipage}
	\begin{minipage}{0.98\linewidth}
		\centering
		\subfigure{\includegraphics[width=\imgwidth]{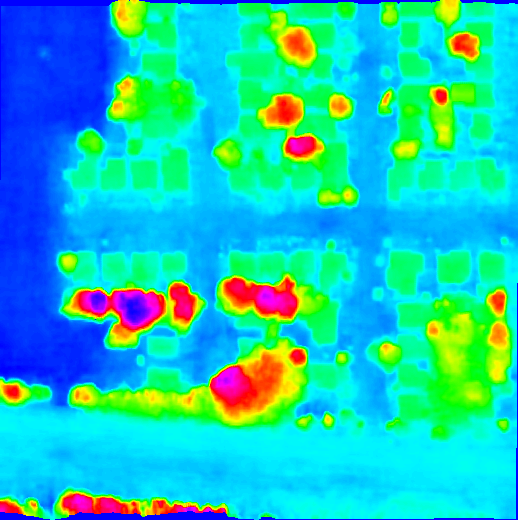}}
		\hfill
		\subfigure{\includegraphics[width=\imgwidth]{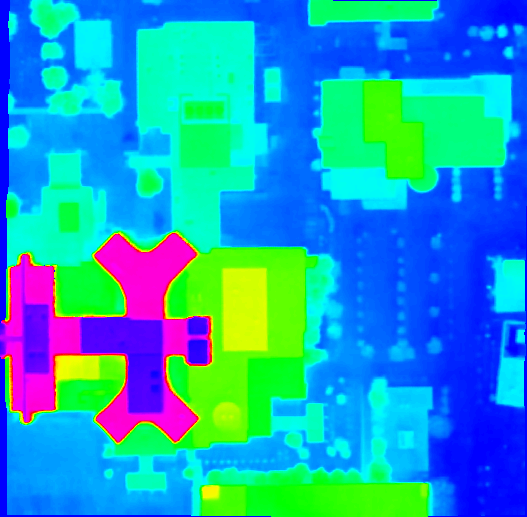}}
		\hfill
		\subfigure{\includegraphics[width=\imgwidth]{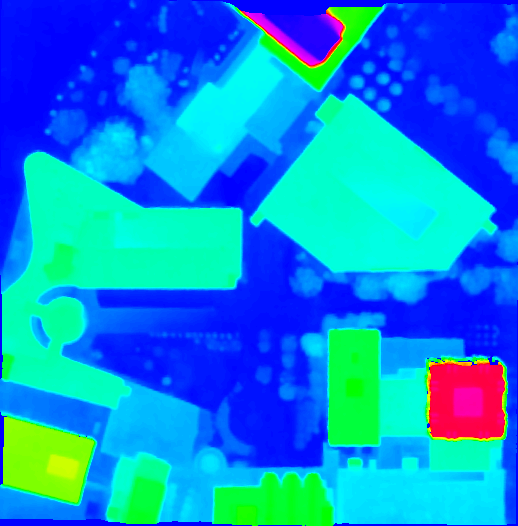}}
		\hfill
		\subfigure{\includegraphics[width=\imgwidth]{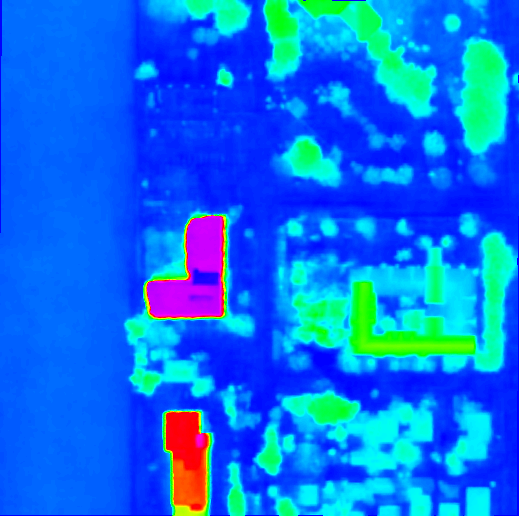}}
	\end{minipage}
	
	\caption{
		Extracted DSM of our proposed regularization in comparison to the Lidar Ground Truth, \emph{EO-NeRF*} \cite{eonerf} and \emph{EO-GS} \cite{eogs}. For each scene the view with minimal MAE is shown.
		Colors map height from blue (low) to red (high).
	}
	\label{fig:results_all}
\end{figure*}

\Cref{fig:results_neighbours} shows that \emph{NeRF}s systematically warp geometry to fit each view, as evidenced by misaligned surface normals. Our proposed \emph{Planarity Regularization} improves surface normal quality by enforcing gravity alignment.
\Cref{fig:results_all} demonstrates that our method produces more accurate scene geometry, effectively mitigating local geometrical artifacts introduced by overfitting.

\subsection{Ablation Study}

We present an ablation study evaluating the impact of our proposed regularization terms (\emph{Geometrical Granularity},
\emph{Gravity\hyp{}Aligned} \emph{Planarity} and \emph{Depth Supervised}) on reconstruction quality in \cref{tab:eval_ablation_study}.
The results demonstrate that introducing individual geometric regularizations improve the MAE compared to the state-of-the-art baselines. Notably, the combination of all three regularization terms achieves the lowest mean MAE, indicating improved geometric accuracy overall. However, we observe that for certain scenes, such as scene 068, the combination of frequency and planar regularization yields the lowest MAE, while the full combination achieves the best result for other scenes such as 260. This suggests that while the regularizations are complementary and their combination provides the most robust performance across all scenes, specific combinations may be more effective for individual cases. 
PSNR values remain relatively stable across different configurations, indicating that improvements in geometry do not come at the expense of image reconstruction quality.

\subsection{Minimizing Computational Cost}
\label{subsec:ablation_studies}

To reduce the computational impact of determining explicit surface normals, we propose reducing the evaluated length of the adjacent rays in \cref{subsubsec:computational_cost}, therefore reducing the required samples which in turn lowers the computational cost. In \cref{tab:ablation_compute} we show that centering adjacent rays around the main surface depth with a length of 50\% only has marginal impacts on reconstruction quality for most scenes, even decreasing the MAE for some such as 068.

\begin{table}[h]
	\begin{center}
		{
			\footnotesize\tabcolsep3pt
			\begin{tabular}{cc|cccc|c}
				\toprule
				& & \multicolumn{5}{c}{\textbf{MAE [m] $\downarrow$}}\\
				Length & Samples &004&068&214&260& Mean\\
				\midrule

				100\% & $2 \times 128$ & 1.30 & 1.08 & 2.13 & 1.39 & 1.48 \\
				50\% & $2 \times 64$ & 1.37 & 1.05 & 2.11 & 1.47 & 1.50 \\
				25\% & $2 \times 32$ & 1.91 & 1.02 & 2.12 & 1.45 & 1.63 \\
				\bottomrule
			\end{tabular}
		}
	\end{center}
	\caption{Ablation study on adjacent ray length and sample density for explicit normal calculation. Centering adjacent rays around main surface point with a length of 50 \% reduces computation with minimal impact on MAE for most scenes. }
	\label{tab:ablation_compute}
\end{table}

\clearpage

\section{Conclusion}
\label{sec:conclusion}
This paper introduces \modelname{} featuring three model\hyp{}agnostic geometry regularization techniques: \emph{Gravity\hyp{}Aligned}\linebreak \emph{Planarity},  \emph{Geometrical Granularity} and \emph{Depth Supervised Regularization}. By aligning local surface approximations with the axis of gravity, we are able to provide geometrical guidance to the \emph{NeRF}. The \emph{Granularity Regularization} limits the available frequencies during training, forcing the network to learn the geometry in a coarse-to-fine manner. 
Lastly we reintroduce \emph{Depth Supervised Regularization}, guiding the network during the initial training stage using a coarse 3D point cloud of the scene.

Our proposed regularization techniques are able to remove high-frequency geometric artifacts caused by overfitting, improving the MAE by a mean of 14.0\% and 11.4\% for the DFC2019 dataset compared to state-of-the-art baselines such as \emph{EO-NeRF} and \emph{EO-GS}.

{
	\begin{spacing}{1.17}
		\normalsize
		\raggedbottom          %
		\balance               %
		\bibliography{geometry_nerf} %
	\end{spacing}
}

\end{document}